\documentclass[10pt,twocolumn,letterpaper]{article}

\usepackage{cvpr}              

\usepackage[dvipsnames]{xcolor}

\usepackage{listings}
\usepackage{multirow}
\usepackage[accsupp]{axessibility}

\definecolor{cvprblue}{rgb}{0.21,0.49,0.74}
\usepackage[pagebackref,breaklinks,colorlinks,citecolor=cvprblue]{hyperref}

\def\paperID{4005} 
\def\confName{CVPR}
\def\confYear{2024}

\title{ScanFormer: Referring Expression Comprehension by Iteratively Scanning}

\author{
    Wei Su\textsuperscript{\rm 1}\quad
    Peihan Miao\textsuperscript{\rm 2}\quad
    Huanzhang Dou\textsuperscript{\rm 1}\quad
    Xi Li\textsuperscript{\rm 1,\rm 3\thanks{corresponding author.}}\\
    \textsuperscript{\rm 1}College of Computer Science and Technology, Zhejiang University\\
    \textsuperscript{\rm 2}School of Software Technology, Zhejiang University\\
    \textsuperscript{\rm 3}Zhejiang-Singapore Innovation and Al Joint Research Lab\\
    {\tt\small \{weisuzju, peihan.miao, hzdou, xilizju\}@zju.edu.cn}
}

\begin{document}
\maketitle
\begin{abstract}
Referring Expression Comprehension (REC) aims to localize the target objects specified by free-form natural language descriptions in images.
While state-of-the-art methods achieve impressive performance, they perform a dense perception of images, which incorporates redundant visual regions unrelated to linguistic queries, leading to additional computational overhead.
This inspires us to explore a question: can we eliminate linguistic-irrelevant redundant visual regions to improve the efficiency of the model?
Existing relevant methods primarily focus on fundamental visual tasks, with limited exploration in vision-language fields.
To address this, we propose a coarse-to-fine iterative perception framework, called ScanFormer.
It can iteratively exploit the image scale pyramid to extract linguistic-relevant visual patches from top to bottom.
In each iteration, irrelevant patches are discarded by our designed informativeness prediction.
Furthermore, we propose a patch selection strategy for discarded patches to accelerate inference.
Experiments on widely used datasets, namely RefCOCO, RefCOCO+, RefCOCOg, and ReferItGame, verify the effectiveness of our method, which can strike a balance between accuracy and efficiency.
\end{abstract}
\section{Introduction} 

As a fundamental task in vision-language understanding, Referring Expression Comprehension (REC) \cite{transvg,qrnet,word2pix,seqtr} relies on free-form natural language descriptions to identify the referred target object. 
The development of REC not only can underpin various vision-language tasks \cite{zhu2016visual7w,jiang2022visual,salvador2016faster,mcn}, but also potentially contribute to real-world applications such as human-computer interaction \cite{yang2019fast,rong2019unambiguous}.

\begin{figure}
    \centering
    \includegraphics[width=0.45\textwidth]{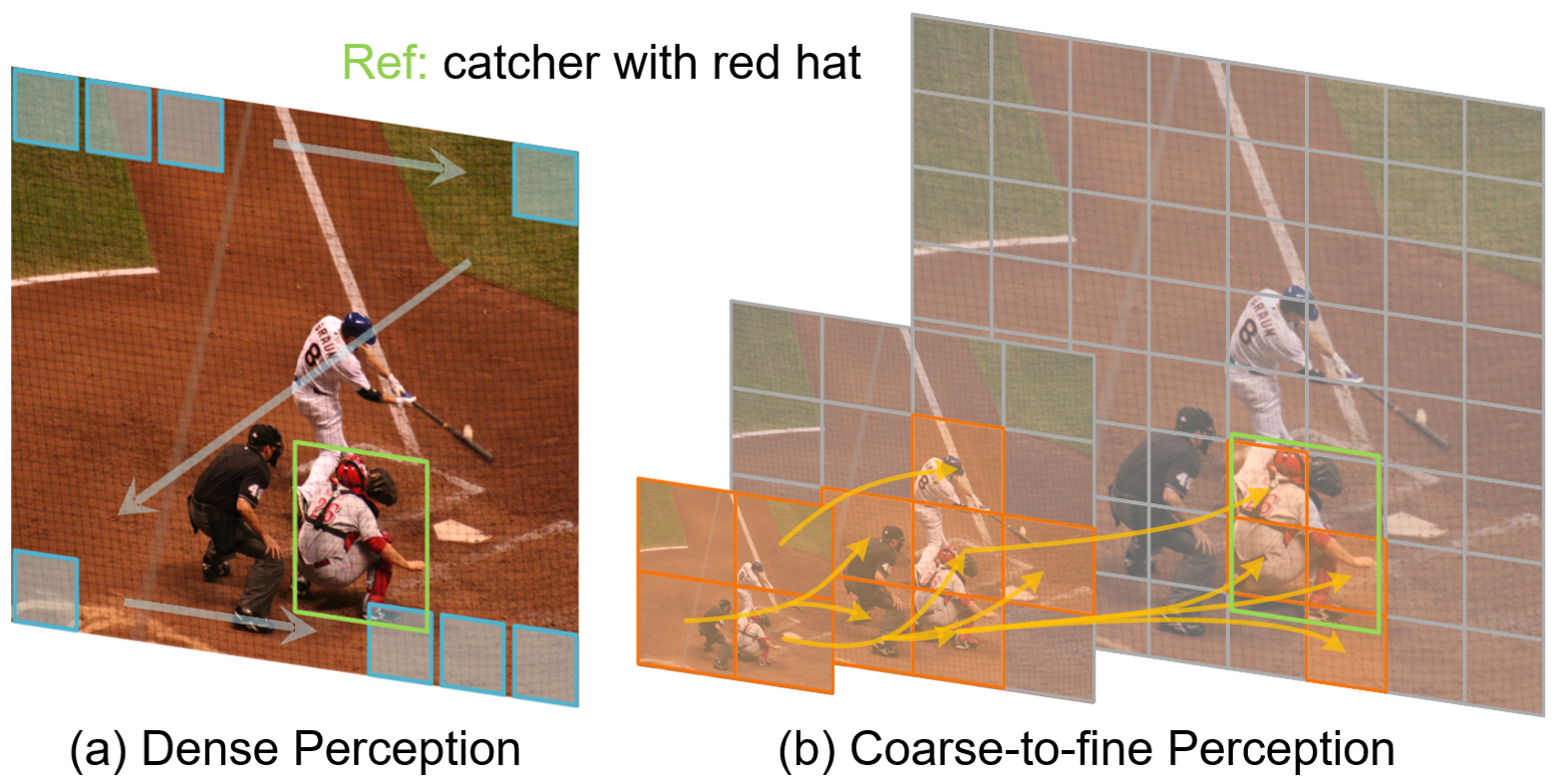}
    \caption{The comparison of dense perception and coarse-to-fine iterative perception. The dense perception extracts features with sliding windows or non-overlapping patches by traversing the image.
    In contrast, our iterative perception can identify and discard linguistic-irrelevant redundant regions from coarse to fine scales.}
    \label{fig:first_fig}
\end{figure}

In REC, images typically contain a substantial amount of redundant information when contrasted with highly concise and information-dense linguistic queries.
For instance, as shown in \cref{fig:first_fig}, the image has considerable redundant visual regions that are weakly correlated or even unrelated to the language query, such as persons around the target catcher and extensive low-information background regions.
However, state-of-the-art methods \cite{transvg,qrnet,word2pix} adopt the form of dense perception to obtain visual features for subsequent cross-modal interaction.
These methods use visual encoders such as ResNet \cite{resnet}, DarkNet \cite{yolov3}, Swin Transformer \cite{swin}, \textit{etc.}, and traverse the entire spatial locations of the image using sliding windows or non-overlapping patches to extract features, as shown in \cref{fig:first_fig} (a).
Despite achieving impressive performance, the form of dense perception brings a significant amount of redundant information and increases computational overhead for the entire model.
Especially in Transformer-based models \cite{transvg,qrnet}, the computational complexity of multi-head self-attention \cite{transformer} is quadratic.
This leads to a research question: \textbf{is it possible to discard linguistic-irrelevant redundant visual regions to enhance the efficiency of the model?}

It is worth noting that there is an emerging trend \cite{dvt,cfvit,tcf,pvt,tome} to explore the elimination of redundant visual features.
Typical bottom-up merging methods \cite{pvt,tcf,tome} initially divide the images into fine-grained patches and gradually merge the patches in subsequent multiple stages to reduce visual tokens.
However, the initial abundance of tokens inevitably leads to a substantial computational cost in the early stages, especially when dealing with high-resolution images.
In addition, the top-down coarse-to-fine methods \cite{dvt,cfvit} start with coarse-grained partitioning using a large patch size, and gradually decrease the patch size to obtain fine-grained visual tokens.
For instance, DVT \cite{dvt} cascades multiple Transformers, and uses confident predictions to determine whether to divide the entire image into finer-grained patches using a smaller patch size.
However, this method usually brings considerable redundant visual regions and increases computational overhead.
CF-ViT \cite{cfvit} introduces a coarse-to-fine two-stage vision Transformer, which identifies informative patches in the coarse stage and further re-split them into finer patches in the second stage.
Although impressive performance in classification, the heuristic informative region identification based on class attention limits its extension to other tasks and models without the [CLS] token.
Furthermore, since it is non-learnable, applying regularization to control token sparsity is challenging.
Therefore, existing efficient Transformer methods still have limitations, and focus on visual tasks while ignoring the exploration of the vision-language fields.

To address this, this paper proposes a coarse-to-fine iterative perception framework, termed \textbf{ScanFormer}, as shown in \cref{fig:first_fig} (b).
To be specific, using a pre-constructed image scale pyramid, the model initiates visual perception from the coarse-grained and low-resolution image at the top of the pyramid.
By predicting the informativeness of finer-grained patches in the next iteration, the model adaptively eliminates redundant visual regions, ultimately reaching the fine-grained and high-resolution image at the bottom of the pyramid.
We keep previous tokens in the cache without further updates, thus reducing computational resources. 
The new tokens extracted in each iteration interact with themselves and previous tokens contained in the cache via self-attention and cross-attention, respectively.
In this process, multi-scale patch partitioning enables the model to aggregate scale-related information from different spatial positions.
Furthermore, we propose a patch selection strategy for discarded patches to accelerate inference.
A learnable token participates in the coarse-to-fine iterative perception process and is ultimately utilized for coordinate regression to directly predict the target box.
Extensive experiments have demonstrated the effectiveness of our ScanFormer, which achieves state-of-the-art methods on widely-used datasets, \textit{i.e.,} RefCOCO \cite{refcoco}, RefCOCO+ \cite{refcoco}, RefCOCOg \cite{refcocog-google}, and ReferItGame \cite{referitgame}.

The main contributions can be summarized as follows:
\begin{itemize}
    \item We propose ScanFormer, a coarse-to-fine iterative perception framework that progressively discards linguistic-irrelevant redundant visual regions in each iteration to enhance the efficiency of the model.
    \item To achieve patch selection, we propose to select tokens by constant token replacement, where the unselected tokens are replaced by a constant token and merged finally for real acceleration.
    \item Extensive experiments demonstrate the effectiveness of our ScanFormer, which strikes a balance between accuracy and efficiency compared to state-of-the-art methods.
\end{itemize}

\section{Related Work}

\subsection{Referring Expression Comprehension}
Most conventional methods \cite{mattnet,rvg-tree,cm-att-erase,ref-nms,ddpn} explore REC through a two-stage framework.
Concretely, in the first stage, numerous candidate proposals for the input image are pre-generated using a pre-trained object detector \cite{faster}.
In the second stage, the proposal that best matches the given referring expression is considered the referred target box.
However, two-stage methods are constrained by the accuracy and speed of the object detector.
To this end, one-stage methods \cite{realgin,mcn,resc,yang2019fast} based on dense anchors \cite{yolov3} are proposed, which can achieve faster speed and comparable performance to the two-stage methods.
In recent years, the success of the transformer \cite{transformer} in vision-language fields has attracted researchers, leading to the emergence of REC methods \cite{qrnet,word2pix,seqtr,transvg,transvg++,lads} based on the transformer.
Due to the multi-head attention mechanism \cite{transformer}, transformer-based REC methods can effectively capture cross-modal relationships.
While achieving impressive performance, these methods incur additional computational overhead due to their dense perception of images.
To this end, this paper proposes a coarse-to-fine iterative perception framework to enhance the efficiency of the model.

\begin{figure*}[t]
    \centering
    \includegraphics[width=\textwidth]{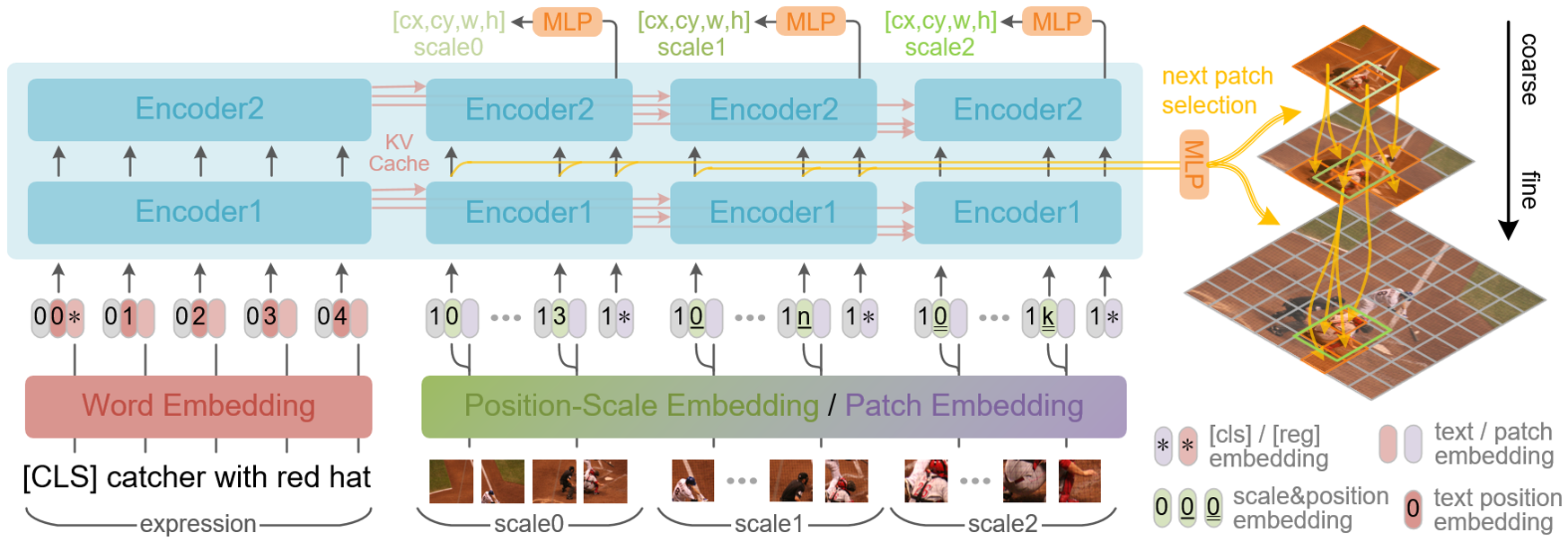}
    \caption{The overall architecture of ScanFormer. The text inputs and image patches at each scale share the encoder. The outputs of the first half part of the encoder, \textit{i.e.} Encoder1, are used to select finer-grained patches for the next level. The [REG] tokens output by the second half of the encoder, \textit{i.e.} Encoder2, are used to predict the coordinates of the referred object at the corresponding scale. The key and value features generated in the encoder are cached and propagated from left to right.}
    \label{fig:main_arch}
\end{figure*}

\subsection{Efficient Vision Transformer}
The self-attention mechanism \cite{transformer} is the primary reason for the inefficiency of ViT \cite{vit}, as its computational complexity grows quadratically with the number of visual tokens.
Recently, several methods \cite{dvt,cfvit,tcf,pvt,tome} have have emerged to improve the efficiency of ViT by reducing the number of computed visual tokens.
These methods can be broadly categorized into bottom-up token merging methods \cite{pvt,tcf,tome} and top-down coarse-to-fine methods \cite{dvt,cfvit}.
To be specific, bottom-up token merging methods \cite{pvt,tcf,tome} initially divide the high-resolution image into fine-grained patches, and gradually merge these patches in multiple stages to reduce the number of visual tokens.
In addition, top-down coarse-to-fine methods \cite{dvt,cfvit} start with coarse-grained partitioning, \textit{i.e.} large patch size but a small number of tokens, and progressively reduce the patch size while performing fine-grained partitioning.
However, existing relevant methods focus on efficient vision Transformers, with a limited exploration into efficient vision-language Transformers.
In this paper, we explore an efficient vision-language Transformer framework for REC.

\section{Method}
In this section, we give a detailed description of our ScanFormer for REC.
First, we briefly introduce the overview of our framework in \cref{method:sec1}.
Then, we elaborate on the patch selection strategy in \cref{method:sec2}.
Next, we describe our prediction head in \cref{method:sec3}.
Finally, we detail the training objectives of the whole framework in \cref{method:sec4}.

\subsection{Framework}
\label{method:sec1}
\begin{figure}
    \centering
    \includegraphics[width=0.4\textwidth]{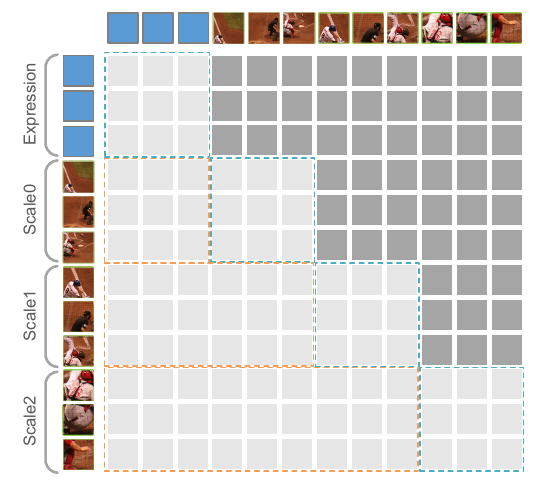}
    \caption{Token interaction of different modalities and scales. "dark" color means blocking interaction. The regions surrounded by blue dotted lines represent the interaction in each iteration, and the regions surrounded by orange dotted lines represent the interaction with the K\&V cache.}
    \label{fig:attn_mask}
\end{figure}

ScanFormer utilizes a unified Transformer-like structure for linguistic and visual modalities, as illustrated in \cref{fig:main_arch}.
Concretely, the framework consists of word embedding, patch embedding, position-scale embedding, and encoders.
Word embedding and patch embedding extract features from texts and images, respectively. 
Position-scale embedding is used to encode the spatial position and scale size of each image patch.
The encoder consists of $N$ layers, each comprising a Multi-Head Attention (MHA) layer and a Feed-Forward Network (FFN).
In addition, each encoder layer is equipped with a cache to store the output features.
The query for MHA comes from the input features, while the key and value are composed of features from the input features and the previous cached features, as illustrated in \cref{fig:attn_mask}.
The causality in scale not only reduces the amount of calculations but also leverages previous linguistic and multi-scale visual information to update features.

The input of linguistic modality is initially encoded by the framework, and the extracted linguistic features are stored in the cache.
Subsequently, for the visual modality, an image scale pyramid with $S$ scales is constructed based on the input image $I$. 
From top to bottom, for each iteration, selected patches are extracted and processed through the framework, where intermediate features are used to generate the selection of sub-patches in the next pyramid layer.
In addition, the cache at each layer of the encoder stores the visual features obtained after each iteration.
The features corresponding to the [REG] token in each iteration are used to predict the coordinates of the referred object at the corresponding scale.
In particular, for the image at the top of the pyramid, all the patches are selected to ensure that the model captures global information.
As the scale increases, ScanFormer incorporates finer-grained features to achieve accurate predictions, while discarding irrelevant patches to save substantial computing resources.

Specifically, for the linguistic modality, the referring text $t\in\mathbb{R}^{L\times|V|}$ is embedded with the word embedding matrix $T\in\mathbb{R}^{|V|\times d}$, prepended the [CLS] embedding $T^{cls}\in\mathbb{R}^d$, and then added with the text position embedding matrix $T^{pos}\in\mathbb{R}^{(L+1)\times d}$ and the type embedding $T^{type}\in\mathbb{R}^d$.
The embedded linguistic features are first fed into the framework, and the updated linguistic features are stored in the cache at each layer of the encoder.
For the visual modality, from top to bottom of the image scale pyramid, taking level-$i$ as an example, the $N_i$ selected patches with $(P, P)$ resolution and $C$ channels are first flattened to $v\in\mathbb{R}^{N_i\times (P^2\cdot C)}$
and then projected to $E\in\mathbb{R}^{N_i \times d}$ with a linear projection layer.
After that, the patch features are added with the spatial embedding $E^{spatial}\in\mathbb{R}^{N_i\times d}$ and the type embedding $E^{type}\in\mathbb{R}^d$. 
$E^{spatial}$ is produced by the position-scale embedding $PSE: [0,1]^3 \rightarrow \mathbb{R}^d$ with the normalized patch coordinates and scales $[cx, cy, s]$ as inputs. 
After that, The embedding of the [REG] token $E^{reg}\in\mathbb{R}^d$ is appended, which is used to regress the bounding box $[cx_i, cy_i, w_i, h_i]$ of the object at level $i$.

\subsection{Patch Selection by Constant Replacement}
\label{method:sec2}
To facilitate learning to select informative patches through back-propagation, a selection factor $s_i$ is generated for the $i$-th patch.
There are two options for using $s_i$:
\textbf{(1)} Apply $s_i$ to every head of the MHA on every Transformer layer. 
This is achieved by weighting the key and value, and gradually decaying $s_i$ to $0.0$ to minimize its impact on the remaining tokens.
However, for a Transformer with $N$ layers and $H$ heads, obtaining clear gradient signals to optimize $s_i$ is challenging, making it difficult to achieve ideal learning choices.
\textbf{(2)} Apply $s_i$ directly to the inputs of the Transformer, \textit{i.e.} patch embedding, in a weighted manner. 
Since $s_i$ is only used at this location, it is easier to train. 
Therefore, this paper adopts the second candidate.

Furthermore, it is worth noting that even if the input patch embedding is set to zero, it still becomes non-zero in subsequent layers due to the bias terms of FFN and MHA, and the dot product attention.
Fortunately, when the token sequence contains many identical tokens, the calculation of MHA can be simplified, leading to practical inference acceleration.
To improve the model's adaptability, the paper suggests replacing the patch embedding with a learnable constant token rather than directly setting it to zero.
Therefore, the patch selection problem is transformed into a patch replacement problem.
Next, the constant patch replacement and merging for acceleration will be introduced.

\paragraph{Constant Token Replacement.}
To implement the token replacement, a constant token $E^{const}\in\mathbb{R}^d$ is introduced and the selection logits $r_i\in \mathbb{R}$ for $i$-th patch is yielded from the Transformer. 
We follow the improved semantic hashing \cite{kaiser2018discrete} to learn $r_i$ by back-propagation. 
To encourage exploration, noises are added to $r_i$, \textit{i.e.} $r_i^n=r_i+n$. During training, $n\sim \mathcal{N}(0,1)$, and $n=0$ when evaluation and inference. Then, two variables $v_1=\sigma{'}(r_i^n)$, and $v_2=\mathbb{I}(r_i^n\geq 0)$ can be calculated.
\begin{equation}
    \sigma{'}(x) = clamp(1.2\sigma(x)-0.1, 0, 1),
\end{equation}
where $\mathbb{I}(\cdot)$ and $\sigma(\cdot)$ are the indicative function and $sigmoid$ respectively.
During training, in the forward pass, we uniformly sample $v_1$ and $v_2$ as the selection factor $s_i$.
\begin{equation}
\label{equ:select}
    s_i = \mathbb{I}(n_s\geq0.5)\cdot v_1 +\mathbb{I}(n_s < 0.5)\cdot v_2,
\end{equation}
where $n_s\sim Uniform[0, 1]$ represents the random sample weight.
In the backward pass, the gradients always flow to $v_1$, even if $v_2$ is used in the forward computation.
The weighted patch embedding $\overline{E}_i$ can be calculated as:
\begin{equation}
    \overline{E}_i = s_i\cdot E_i + (1-s_i)\cdot E^{const}.
\end{equation}
During training, $s_i$ is regularized to $0$, \textit{i.e.} the $i$-th token is replaced by the constant token $E^{const}$.

\paragraph{Merging Constant Tokens.}
Although the redundant tokens are replaced by the constant tokens and are still included in the forward computation of the encoder, they can not be discarded directly without any impact. 
However, it can be proved that these constant tokens can be merged to reduce the computation effectively. 
Taking a key and value sequences with $N$ tokens and $N_c$ constant tokens:
\begin{equation}
\label{equ:orig_kv}
    \begin{split}
        &K=[\underbrace{k_1, k_2, \cdots, k_i}_{N-N_c}, \underbrace{k^c,\cdots, k^c}_{N_c}] \\
        &V=[\underbrace{v_1, v_2, \cdots, v_i}_{N-N_c}, \underbrace{v^c,\cdots, v^c}_{N_c}]
    \end{split}
\end{equation}
The keys and values of $N_c$ tokens can be reduced to only one key and value by concatenating a constant vector to keys, which can be illustrated as: 
\begin{equation}
\label{equ:merge_kv}
    \begin{split}
        &K^{'}=concat([\underbrace{k_1, k_2, \cdots, k_i}_{N-N_c}, k^c], [\underbrace{0, 0, \cdots, 0}_{N-N_c}, log(N_c)]) \\
        &V^{'}=[\underbrace{v_1, v_2, \cdots, v_i}_{N-N_c}, v^c]
    \end{split}
\end{equation}
According to the scaled dot-product attention mechanism, the attention values $A\in\mathbb{R}^{N}$ for one query $q\in\mathbb{R}^{d}$ relative to $K$ can be calculated as:
\begin{equation}
\label{equ:attn}
A = softmax(\frac{qK^T}{\sqrt{d}}).
\end{equation}
It can be concluded that the same attention-weighted value can be derived using \cref{equ:orig_kv} and \cref{equ:merge_kv} according to \cref{equ:attn}. Therefore, $N_c-1$ tokens are eventually dropped and the computations brought by them can be saved. The Pytorch \cite{pytorch} implementation is illustrated in Algorithm \ref{alg_attn_merge}.
\renewcommand{\lstlistingname}{Algorithm}
\begin{lstlisting}[caption=Dot-product Attention with Constant Tokens, label=alg_attn_merge]
"""
The features of constant tokens are stored to 
the last element of key and value.
Input:
  query: (bsz, dst_len, ndim)
  key, value: (bsz, src_len, ndim)
  num_const: (bsz, 1)
Output:
  attn_out: (bsz, dst_len, ndim)
"""
pad_query = torch.ones(bsz, dst_len, 1)
pad_key = torch.zeros(bsz, src_len, 1)
pad_key[:, -1] = torch.log(num_const)

q = torch.cat([query, pad_query], dim=-1)
k = torch.cat([key, pad_key], dim=-1)

attn = torch.softmax(q@k.t(-1, -2), dim=-1)
attn_out = attn @ value
\end{lstlisting}

\subsection{Prediction Head}
\label{method:sec3}
The referred object may exist at various scales. Similar to the object detection methods\cite{faster,yolov3}, where multi-scale predictions are conducted at different feature levels, for each scale level in ScanFormer, we apply direct coordinate regression \cite{transvg} to predict the bounding box of the referred object. 
The regression token [REG] is introduced to gather features of image patches across the Transformer.
The output features corresponding to the [REG] token is fed to a shared multi-layer perception (MLP), followed by the Sigmoid function to predict the normalized bounding box $\hat{b}=(\hat{x},\hat{y},\hat{w},\hat{h})$ of the referred objects.

\subsection{Training Objectives}
\label{method:sec4}
We optimize the proposed coarse-to-fine iterative perception framework end-to-end.
For the $l$-th image scale, we can obtain predicted bounding box $\hat{b}_l=(\hat{x}_l,\hat{y}_l,\hat{w}_l,\hat{h}_l)$.
Given the ground truth $b=(x,y,w,h)$, the detection loss function is defined as follows:
\begin{equation}
\label{equ:loss_bbox}
    \mathcal{L}_{bbox} = \sum_{l=0}^{2} \lambda_{L1}^l\mathcal{L}_{L1}(b,\hat{b}_l) + \sum_{l=0}^{2} \lambda_{giou}^l\mathcal{L}_{giou}^l(b,\hat{b}_l),
\end{equation}
where $\mathcal{L}_{L1}(\cdot,\cdot)$ and $\mathcal{L}_{giou}(\cdot,\cdot)$ represent L1 loss and Generalized IoU loss \cite{giou}, respectively, and $\lambda_{L1}^l$ and $\lambda_{giou}^l$ are the relative weights to control the detection loss penalty for the $l$-th image scale.

In addition, to control the sparsity of selected patches, we add a regularization loss function as follows:
\begin{equation}
\label{equ:loss_sparse}
    \mathcal{L}_{sparse} = \lambda_{sparse}\sum_{l=1}^{2}(\frac{1}{N_l}\sum_{i=1}^{N_l}s_i^l-\beta^l)^2,
\end{equation}
where $\lambda_{sparse}$ represents the relative weights to control the sparsification penalty, and $s_i^l$ represents the selection factor for the $i$-th patch in \cref{equ:select} in the $l$-th image scale. $\beta^l$ is the hyperparameter to control the ratio of selected tokens from the $l$-th image scale.
The total loss function of our ScanFormer is defined as follows:
\begin{equation}
    \mathcal{L}_{total} = \mathcal{L}_{bbox} + \mathcal{L}_{sparse},
\end{equation}
The trained ScanFormer can strike a balance between accuracy and efficiency.
The experimental analysis of the ScanFormer will be elaborated in \cref{experiments}.

\section{Experiment} 
\label{experiments}

In this section, we provide a detailed experimental analysis of the entire framework, including the datasets, evaluation protocol, training and inference implementation details, comparisons with state-of-the-art methods, early exiting, and qualitative results.

\subsection{Datasets and Evaluation Protocol}

\paragraph{Datasets.}
To demonstrate the effectiveness of our method, we conduct experiments on the widely used REC dataset, which includes RefCOCO \cite{refcoco}, RefCOCO+ \cite{refcoco}, RefCOCOg \cite{refcocog-umd}, and ReferItGame \cite{referitgame}.
RefCOCO, RefCOCO+ and RefCOCOg are constructed based on MS-COCO \cite{mscoco}.
To be specific, RefCOCO and RefCOCO+ are collected from interactive games, including train, val, testA, and testB sets.
In contrast to RefCOCO, expressions in RefCOCO+ do not contain words related to the absolute position of the referred objects.
Unlike RefCOCO and RefCOCO+, RefCOCOg is collected on Amazon Mechanical Turk in a non-interactive setting, which results in longer and more complex referring expressions.
Following the common split version \cite{refcocog-umd}, RefCOCOg consists of train, val, and test sets.
In addition, ReferItGame is constructed based on SAIAPR-12 \cite{saiapr12}, including train and test sets.
We also pre-train ScanFormer with a large-scale pre-training dataset, which contains 174k images with approximately 6.1M distinct referring expressions by combining the train sets of RefCOCO/+/g, ReferItGame, Visual Genome regions \cite{vg}, and Flickr entities \cite{flickr30k}.

\begin{table*}[t]
  \footnotesize
  \centering
    \begin{tabular}{c|c|c|ccc|ccc|cc|c}
    \toprule
    \multirow{2}[2]{*}{Methods} & \multirow{2}[2]{*}{Venue} & \multirow{2}[2]{*}{Backbone} & \multicolumn{3}{c|}{RefCOCO} & \multicolumn{3}{c|}{RefCOCO+} & \multicolumn{2}{c|}{RefCOCOg} & ReferItGame \\
          &       &       & val   & testA & testB & val   & testA & testB & val   & test  & test \\
    \midrule
    \textbf{Two-stage:} &       &       &       &       &       &       &       &       &       &       &  \\
    MAttNet\cite{mattnet} & CVPR18 &   ResNet-101/LSTM    & 76.65 & 81.14 & 69.99 & 65.33 & 71.62 & 56.02 & 66.58 & 67.27 & 29.04 \\
    RvG-Tree \cite{rvg-tree} & TPAMI19 & ResNet-101/LSTM & 75.06 & 78.61 & 69.85 & 63.51 & 67.45 & 56.66 & 66.95 & 66.51 & - \\
    CM-Att-Erase \cite{cm-att-erase} & CVPR19 & ResNet-101/LSTM & 78.35 & 83.14 & 71.32 & 68.09 & 73.65 & 58.03 & 67.99 & 68.67 & - \\
    Ref-NMS\cite{ref-nms} & AAAI21 &   ResNet-101/GRU    & 80.70 & 84.00 & 76.04 & 68.25 & 73.68 & 59.42 & 70.55 & 70.55 & - \\
    \midrule
    \textbf{One-stage:} &       &       &       &       &       &       &       &       &       &       &  \\
    FAOA \cite{yang2019fast} & ICCV19 & DarkNet-53/BERT & 72.54 & 74.35 & 68.50  & 56.81 & 60.23 & 49.60   & 61.33 & 60.36 & 60.67 \\
    ReSC-Large \cite{resc}  & ECCV20 & DarkNet-53/BERT & 77.63 & 80.45 & 72.30  & 63.59 & 68.36 & 56.81  & 67.30  & 67.20  & 64.60 \\
    MCN \cite{mcn}  & CVPR20 & DarkNet-53/GRU & 80.08 & 82.29 & 74.98 & 67.16 & 72.86 & 57.31    & 66.46 & 66.01 & - \\
    RealGIN \cite{realgin} & TNNLS21 & DarkNet-53/GRU & 77.25 & 78.70  & 72.10  & 62.78 & 67.17 & 54.21 & 62.75 & 62.33 & - \\
    PLV-FPN \cite{plv} & TIP22 & ResNet-101/BERT & 81.93 & 84.99 & 76.25 & 71.20  & 77.40  & 61.08 & 70.45 & 71.08 & 71.77 \\
    \midrule
    \textbf{Transformer-based:} &       &       &       &       &       &       &       &       &       &       &  \\
    TransVG \cite{transvg} & ICCV21 & ResNet-101/BERT & 81.02 & 82.72 & 78.35 & 64.82 & 70.70  & 56.94 & 68.67 & 67.73 & 70.73 \\
    RefTR \cite{referring-transformer} & NeurIPS21 & ResNet-101/BERT & 82.23 & 85.59 & 76.57 & 71.58 & 75.96 & 62.16      & 69.41 & 69.40  & 71.42 \\
    PFOS \cite{pfos} & TMM22 & ResNet-101/BERT & 78.44 & 81.94 & 73.61 & 65.86 & 72.43 & 55.26 & 67.89 & 67.63  & 67.90 \\
    Word2Pix \cite{word2pix} & TNNLS22 & ResNet-101/BERT & 81.20 & 84.39 & 78.12 & 69.74  & 76.11  & 61.24 & 70.81 & 71.34 & - \\
    SeqTR \cite{seqtr} & ECCV22 & DarkNet-53/GRU & 81.23 & 85.00 & 76.08 & 68.82  & 75.37  & 58.78 & 71.35 & 71.58 & 69.66 \\
    QRNet \cite{qrnet} & CVPR22 & Swin-S/BERT & 84.01 & 85.85 & 82.34 & 72.94  & 76.17  & 63.81 &  71.89 & 73.03 & 74.61 \\
    M-DGT \cite{mdgt} & CVPR22 & ResNet-101/BERT & 85.37 & 84.82 & 87.11 & 70.02 & 72.26 & 68.92 & 79.21 & 79.06  & -\\
    LADS \cite{lads} & AAAI23 & ResNet-50/BERT & 82.85 & 86.67 & 78.57 & 71.16  & 77.64  & 59.82 &  71.56 & 71.66 & 71.08 \\
    \midrule
    \textbf{Ours:} &       &       &       &       &       &       &       &       &       &       &  \\
    ScanFormer & - &  Unified Transformer  & 83.40 & 85.86 & 78.81 & 72.96 & 77.57 & 62.50 & 74.10 & 74.14 & 68.85 \\
    \bottomrule
    \end{tabular}%
  \caption{Comparison with state-of-the-art methods on RefCOCO \cite{refcoco}, RefCOCO+ \cite{refcoco}, RefCOCOg \cite{refcocog-umd} and ReferItGame \cite{referitgame}.}
  \label{tab:sota}%
\end{table*}%

\begin{table}[htbp]
  \footnotesize
  \centering
    \begin{tabular}{c|c@{\hspace{1.0em}}c|c@{\hspace{1.0em}}c|c}
    \toprule
    \multirow{2}[2]{*}{Method} & \multicolumn{2}{c|}{RefCOCO} & \multicolumn{2}{c|}{RefCOCO+} & RefCOCOg \\
          & testA & testB & testA & testB & test\\
    \midrule
    ViLBERT \cite{vilbert} & -     & -     & 78.52  & 62.61  & -  \\
    UNITER\_L \cite{uniter} & 87.04  & 74.17 & 81.45  & 66.70 & 75.77 \\
    VILLA\_L \cite{villa} & 87.48 & 74.84  & 81.54 & 66.84 & 76.71 \\
    MDETR \cite{mdetr} & 89.58 & 81.41 & 84.09 & 70.62 & 80.89 \\
    OFA\_B \cite{ofa} & 90.67 & 83.30 & 87.15 & 74.29 & 82.31 \\
    \midrule
    ScanFormer & 89.99 & 82.89 &  84.04 & 70.63 & 82.75\\
    \bottomrule
    \end{tabular}%
  \caption{Comparison with the large-scale pre-training methods.}
  \label{tab:pretrain}%
\end{table}%

\paragraph{Evaluation Protocol.}
Following the previous works \cite{qrnet,seqtr,transvg}, we choose \emph{Acc}@0.5 as the metric to evaluate the accuracy of positioning the referred objects, where \emph{Acc}@0.5 represents the percentage of predicted correct samples among all test samples.
For each sample, if the intersection-over-union (IoU) between the predicted bounding box and the ground truth is greater than 0.5, it indicates that the predicted bounding box is correct.

\subsection{Implementation Details}

\begin{figure}
    \centering
    \includegraphics[width=0.4\textwidth]{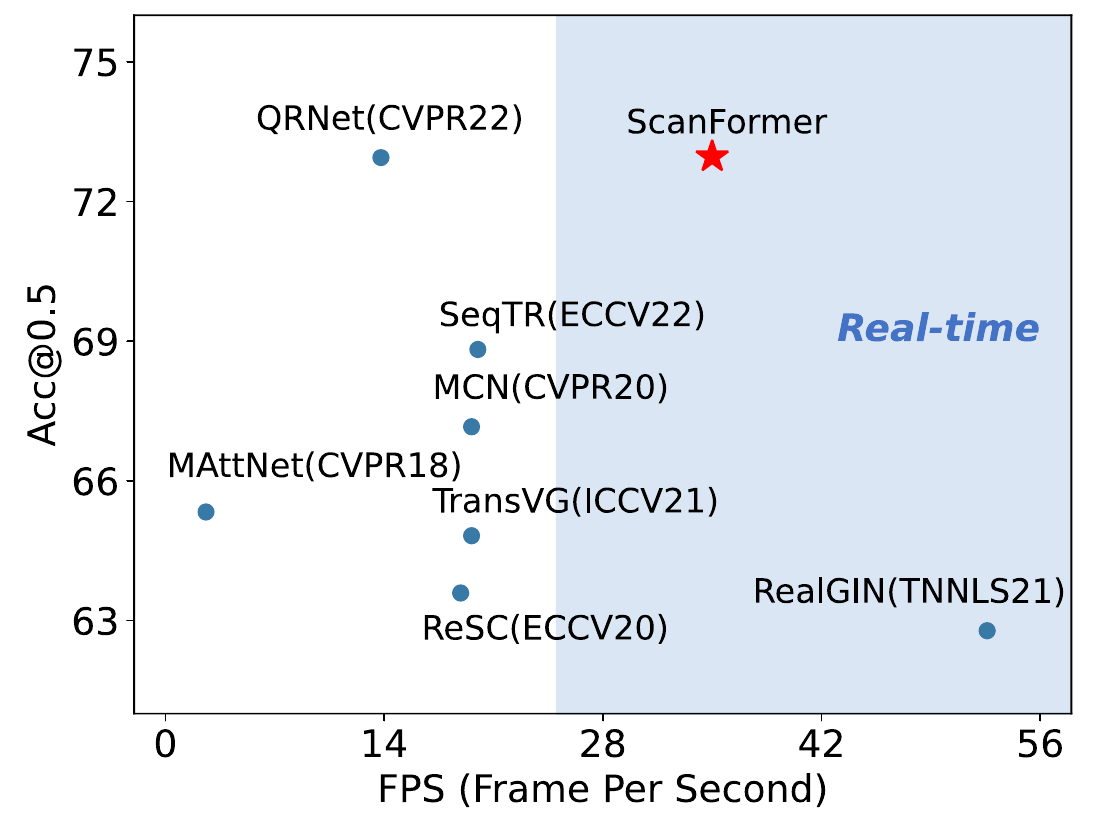}
    \caption{Comparison of the performance and inference speed on the val set of RefCOCO+ \cite{refcoco}. The real-time speed threshold is set to 25 FPS and all inference speeds all tested on the 1080 Ti.}
    \label{fig:fps}
\end{figure}

\paragraph{Training.}
Unlike conventional methods \cite{transvg,qrnet,seqtr} that require additional visual encoder (such as ResNet \cite{resnet}, Swin \cite{swin}) and linguistic encoder (such as LSTM \cite{lstm}, BERT \cite{bert}), the proposed model extract visual and linguistic features using one unified Transformer \cite{transformer}, which is initialized from the ViLT \cite{vilt} pre-training weights.
The resolution of the input image is resized to $640 \times 640$, and the referring expressions are truncated or padded to a length of 40.
Data augmentation operations during training include random color space jittering, Gaussian blur, random horizontal flipping, random cropping, and random resizing.
We set $\lambda_{L1}^l=\lambda_{giou}^l=4^{l-2}$ in \cref{equ:loss_bbox}, and $\lambda_{sparse}=0.05$ and $\beta^l=2^{-l}$ in \cref{equ:loss_sparse}.
The model is optimized end-to-end for 80 epochs using AdamW \cite{adamw}, with a batch size of 384, and weight decay set to $1e^{-4}$.
The learning rate is gradually increased to $1.5e^{-4}$ in the first 800 iterations using a warm-up strategy, and then the learning rate is decayed with a linear strategy.
To test the performance improvement brought by large-scale pre-training, we also pre-train ScanFormer for 40 epochs on the large-scale pre-training dataset and then fine-tune the pre-trained model on the specific data set for 20 epochs.
We implement the framework using PyTorch and conduct experiments using NVIDIA A100 GPUs.

\paragraph{Inference.}
In the inference stage, each input sample consists of an image and a referring expression, where the image is resized to $640 \times 640$, and the maximum length of the referring expression is 40.
Our framework can directly output the bounding boxes specified by referring expressions without any post-processing operations.

\begin{figure}
    \centering
    \includegraphics[width=0.4\textwidth]{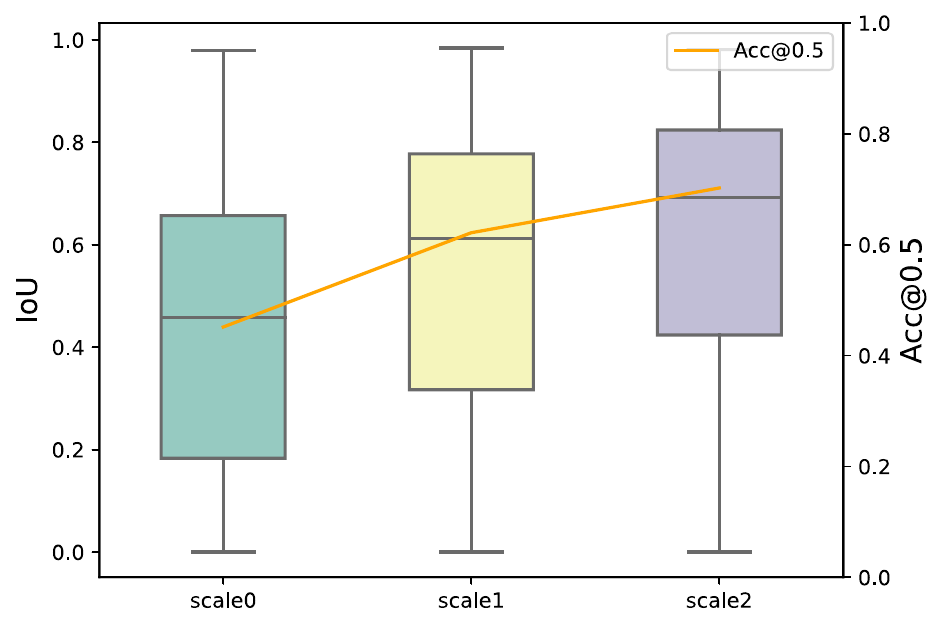}
    \caption{The Acc@0.5 and IoUs between predicted bounding boxes and ground truth of three scales, which are evaluated on the val set of RefCOCOg\cite{refcocog-umd}.}
    \label{fig:iou_acc}
\end{figure}

\subsection{Comparisons with State-of-the-art Methods}
\label{experiments:sec3}
To verify the effectiveness of the ScanFormer proposed in this paper, we compare with state-of-the-art methods on widely used datasets, \textit{i.e.,} RefCOCO \cite{refcoco}, RefCOCO+ \cite{refcoco}, RefCOCOg \cite{refcocog}, and ReferItGame \cite{referitgame}.

Concretely, we compare the performance of our ScanFormer with state-of-the-art REC methods, including two-stage methods \cite{mattnet,rvg-tree,cm-att-erase,ref-nms}, one-stage methods \cite{yang2019fast,resc,mcn,realgin,plv}, and transformer-based methods \cite{transvg,referring-transformer,pfos,word2pix,seqtr,qrnet,mdgt,lads}.
The comparison results are shown in \cref{tab:sota}.
It can be observed that our ScanFormer achieves a significant performance improvement compared to state-of-the-art one-stage method PLV-FPN \cite{plv} and two-stage method Ref-NMS \cite{ref-nms}.
Compared to state-of-the-art transformer-based methods, such as LADS \cite{lads}, SeqTR \cite{seqtr}, and Word2Pix \cite{word2pix}, it can be found that the proposed ScanFormer also achieved good performance.
In particular, compared to QRNet \cite{qrnet}, our ScanFormer achieves comparable performance.
In addition, unlike previous methods that use additional visual and linguistic backbones \cite{resnet,swin,lstm,bert}, ScanFormer only utilizes a unified Transformer to achieve accurate language-to-vision localization.

We also compare ScanFormer with state-of-the-art large-scale pre-training methods, \textit{i.e.} MDETR \cite{mdetr} and OFA \cite{ofa}, as shown in \cref{tab:pretrain}.
Compared with training directly on specific datasets, large-scale pre-training greatly improves the performance of the model.
In addition, ScanFormer achieves comparable performance to MDETR and OFA, but the unified Transformer structure is simpler.

Furthermore, we compare the performance and inference speed of state-of-the-art methods with our ScanFormer on the RefCOCO+ val set, as shown in \cref{fig:fps}.
The inference speed is tested on 1080 Ti, and the proposed ScanFormer achieves a real-time inference speed of 34.9 FPS. 
Compared with state-of-the-art methods \cite{qrnet,transvg,seqtr,mcn,realgin}, we achieve high accuracy and fast inference speed, benefiting from the unified Transformer.
In particular, compared to QRNet \cite{qrnet} with comparable performance, Scanformer achieves an inference speed of more than twice as fast.

\begin{figure}
    \centering
    \includegraphics[width=0.47\textwidth]{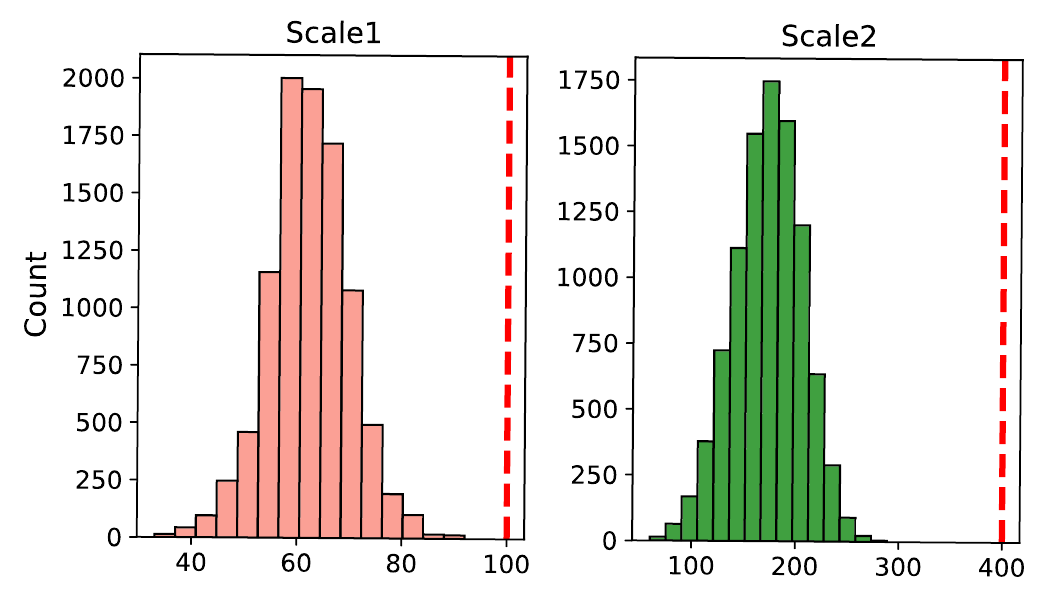}
    \caption{Distribution of the number of selected tokens from different scales in RefCOCOg \cite{refcocog-umd}. The red dotted line indicates the total number of tokens for the corresponding scale.}
    \label{fig:select_stat}
\end{figure}

\subsection{Early Exit in Image Pyramid}
\label{experiments:sec4}
Considering that our ScanFormer iteratively conducts visual perception from coarse granularity to fine granularity, we show the prediction results of the model at different scales, as shown in Figure \cref{fig:iou_acc}.
It can be observed that as the scale level increases, the performance of the model improves significantly, and better bounding box location can be predicted, as shown by the increasing IoU (Intersection over Union) values.
The model can achieve acceptable performance even at the smallest scale.
We also presented the distribution of IoU between predicted boxes and ground-truth boxes at different scales.
It can be found that as the number of iterations increases, the upper and lower bounds of the iou distribution are significantly improved.

Although the gradually improved performance, it is not trivial to select a proper metric to decide when to stop iteration across the scale pyramids like \cite{dvt}.
We made a preliminary attempt by adding an extra branch sibling to the regression head to predict the exit metrics, \textit{e.g.} IoU, GIoU, or the variance of the predicted bounding box \cite{bboxkl}.
The experimental results are not ideal, where the predicted exit metrics have a poor correlation with localization accuracy.
So we left the early exit metric for further exploration.

\subsection{Qualitative Results}
\label{experiments:sec5}

\begin{figure}
    \centering
    \includegraphics[width=0.45\textwidth]{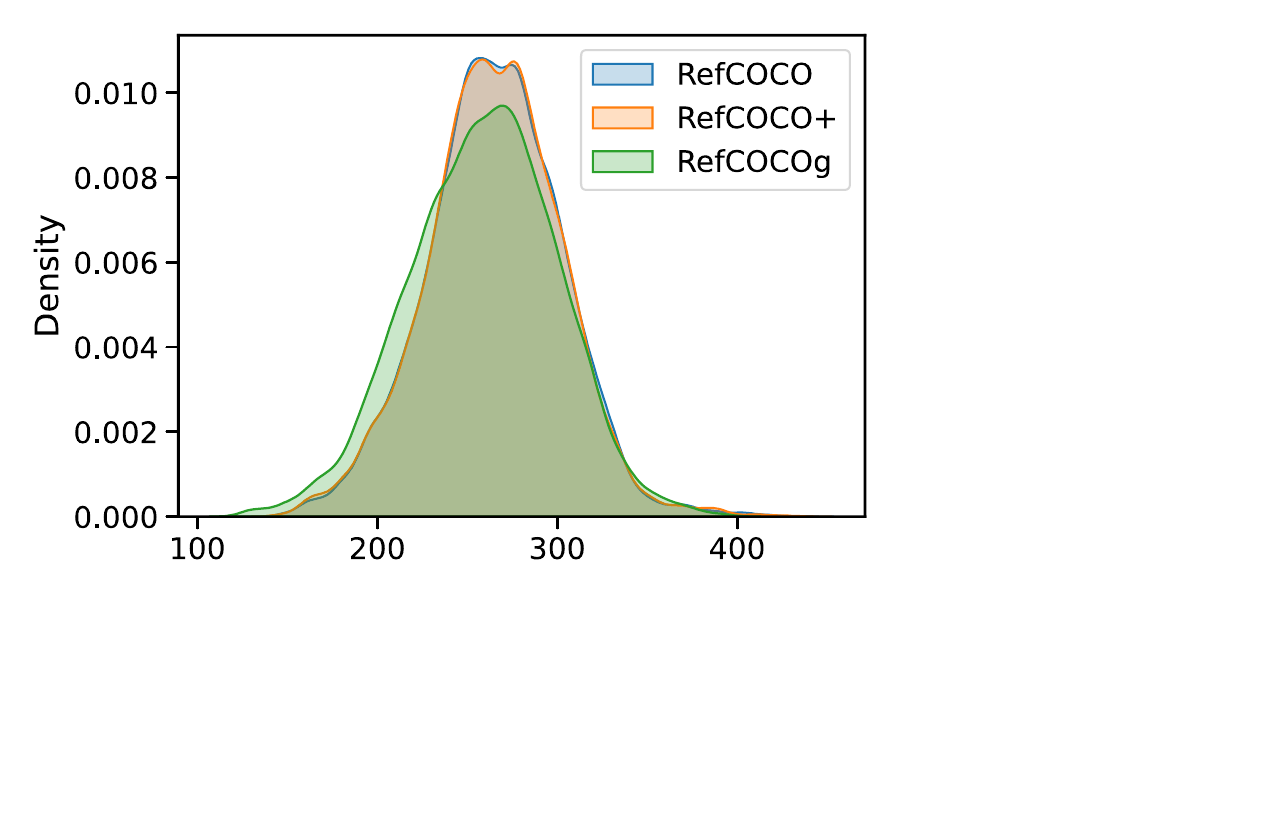}
    \caption{Distribution of the number of selected tokens for samples from RefCOCO \cite{refcoco}, RefCOCO+ \cite{refcoco}, and RefCOCOg \cite{refcocog-umd}.}
    \label{fig:select_stat_sample}
\end{figure}

We propose to reduce computational overhead by replacing not-selected tokens with constant tokens and then merging them.
According to \cref{fig:select_stat}, massive tokens are merged in Scale 1 and Scale 2. The distributions in Scale 0 are not visualized as all the tokens are selected.
There are 40 and 220 tokens replaced with constant tokens on average in Scale 1 and Scale 2, respectively. Token merging can remove 39 and 219 tokens by merging them into one. 
Therefore, token merging can increase speed and significantly reduce FLOPs.
We also visualize the distributions of the number of selected tokens per sample in \cref{fig:select_stat_sample}.
For some samples in RefCOCOg \cite{refcocog-umd}, only 100 tokens are selected for predicting the localization.
Relative to all 400 tokens, each sample has an average of 270 selected tokens participating in the calculation. In addition, causal attention across scales can further reduce computational overhead.

\begin{figure}
    \centering
    \includegraphics[width=0.47\textwidth]{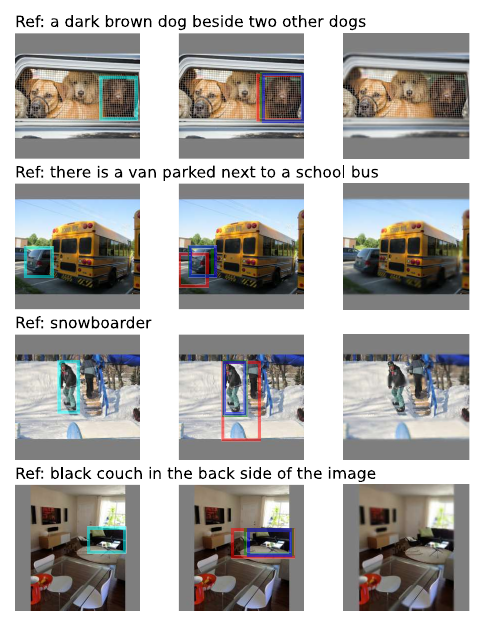}
    \caption{Visualization of examples from the val set of RefCOCOg \cite{refcocog-umd}. From left to right: the input image and ground truth bounding box, the detection results at three scales (red, green, and blue represent the result from scale 0, 1, and 2, respectively), and the image reconstructed from selected patches from different scales.}
    \label{fig:examples}
\end{figure}

The qualitative results are shown in \cref{fig:examples}. 
It can be observed that our model can successfully locate the referred objects, and the regressed bounding boxes are gradually refined with the increasing scales. Furthermore, based on images reconstructed from selected patches, the model can leave regions with low texture or no relation to the reference representation at coarse scales.

\section{Conclusions and Liminations}

In this paper, we explore an efficient vision-language Transformer and propose a coarse-to-fine iterative perception framework, called ScanFormer.
It can continuously discard linguistic-irrelevant redundant visual regions in each iteration to enhance the efficiency of the model.
Extensive experiments on widely used datasets verify the effectiveness of our ScanFormer, which can strike a balance between accuracy and efficiency.
The limitations of our method are two-fold: 
(1) The current method localizes the referred objects through all the scales, and a flexible early exit method can be studied to further improve model efficiency.
(2) The current framework only predicts one target object at a time, limiting its extension to phrase grounding.

\section{Acknowledgments}
This work is supported in part by National Natural Science Foundation of China under Grant U20A20222, National Science Foundation for Distinguished Young Scholars under Grant 62225605, CCF-Zhipu AI Large Model Fund (CCF-Zhipu202302), Zhejiang Key Research and Development Program under Grant 2023C03196, Zhejiang Provincial Natural Science Foundation of China under Grant LD24F020016, SupreMind, and The Ng Teng Fong Charitable Foundation in the form of ZJU-SUTD IDEA Grant, 188170-11102.

{
    \small
    \bibliographystyle{ieeenat_fullname}
    \bibliography{main}
}

\end{document}